\documentclass[11pt]{article}

\usepackage[preprint]{acl}

\usepackage{times}
\usepackage{latexsym}

\usepackage[T1]{fontenc}

\usepackage[utf8]{inputenc}

\usepackage{microtype}

\usepackage{inconsolata}

\usepackage{graphicx}

\usepackage{booktabs}
\usepackage{adjustbox}

\usepackage{hyperref}
\usepackage{graphicx}

\usepackage{placeins} 
\usepackage{enumitem}
\usepackage{minted}
\usepackage{booktabs}

\usepackage{amsmath}%
\usepackage{MnSymbol}%
\usepackage{wasysym}%
\usepackage{xspace}

\usepackage{listings}
\usepackage{xcolor}
\usepackage{minted}
\usepackage{verbatimbox}
\usepackage{alltt}

\usepackage{tabularray}
\usepackage{fontawesome5}

\setlist[itemize]{topsep=0.5pt}

\usepackage[table]{xcolor}
\definecolor{okabe_red}{HTML}{D55E00}
\definecolor{okabe_blue}{HTML}{0072B2}
\definecolor{okabe_green}{HTML}{009E73}
\definecolor{okabe_orange}{HTML}{E69F00}
\definecolor{okabe_purple}{HTML}{CC79A7}
\definecolor{tab_blue}{HTML}{3182BD}
\definecolor{tab_blue2}{HTML}{6BAED6}
\definecolor{tab_blue3}{HTML}{9ECAE1}
\definecolor{tab_orange}{HTML}{E6550D}
\definecolor{tab_orange2}{HTML}{FD8D3C}
\definecolor{tab_orange3}{HTML}{FDAE6B}
\definecolor{tab_green}{HTML}{31A354}
\definecolor{tab_green2}{HTML}{74C476}
\definecolor{tab_green3}{HTML}{A1D99B}
\definecolor{tab_purple}{HTML}{756BB1}
\definecolor{tab_gray}{HTML}{636363}
\definecolor{tab_lightblue}{HTML}{EAF2F8}
\definecolor{tab_lightorange}{HTML}{FCEEE6}
\definecolor{tab_lightgreen}{HTML}{EAF6EE}
\definecolor{tab_lightgray}{HTML}{E0E0E0}

\definecolor{codegray}{gray}{0.95}
\definecolor{codepurple}{rgb}{0.6,0.1,0.1}
\definecolor{codegreen}{rgb}{0,0.5,0}

\lstdefinestyle{mypython}{
    language=Python,
    basicstyle=\ttfamily\scriptsize,
    breaklines=true,
    frame=lines,
    framesep=2mm,
    backgroundcolor=\color{white},
    showstringspaces=false,
    columns=fullflexible,
    keepspaces=true,
    keywordstyle=\color{blue},
    commentstyle=\color{codegreen},
    stringstyle=\color{codepurple},
}

\newcommand{\sequentialitem}{\textcolor{tab_orange}{\textbf{\rule{0.9ex}{0.9ex}}}\,sequential\xspace}

\newcommand{\batteryitem}{\textcolor{tab_orange2}{\textbf{\rule{0.9ex}{0.9ex}}}\,battery\xspace}

\newcommand{\singleitem}{\textcolor{tab_orange3}{\textbf{\rule{0.9ex}{0.9ex}}}\,single-item\xspace}

\usepackage[many]{tcolorbox}

 \title{\texttt{QSTN:} A Modular Framework for Robust Questionnaire Inference\\with Large Language Models}

\author{
  \textbf{Maximilian Kreutner\textsuperscript{1}},
  \textbf{Jens Rupprecht\textsuperscript{1}},
  \textbf{Georg Ahnert\textsuperscript{1}},\\
  \textbf{Ahmed Salem\textsuperscript{1}},
  \textbf{Markus Strohmaier\textsuperscript{1,2,3}}
\\
  \textsuperscript{1}University of Mannheim,
  \textsuperscript{2}GESIS - Leibniz Institute for the Social Sciences,
  \textsuperscript{3}CSH Vienna
}

\begin{document}
\maketitle
\begin{abstract}
We introduce \texttt{QSTN}, an open-source Python framework for systematically generating responses from questionnaire-style prompts to support in-silico surveys and annotation tasks with large language models (LLMs). QSTN enables robust evaluation of questionnaire presentation, prompt perturbations, and response generation methods. Our extensive evaluation (>40 million survey responses) shows that question structure and response generation methods have a significant impact on the alignment of generated survey responses with human answers. We also find that answers can be obtained for a fraction of the compute cost, by changing the presentation method. In addition, we offer a no-code user interface that allows researchers to set up robust experiments with LLMs \emph{without coding knowledge}. We hope that \texttt{QSTN} will support the reproducibility and reliability of LLM-based research in the future. 
\end{abstract}

\section{Introduction}

\begin{figure}[ht]
    \centering
    \includegraphics[width=\linewidth]{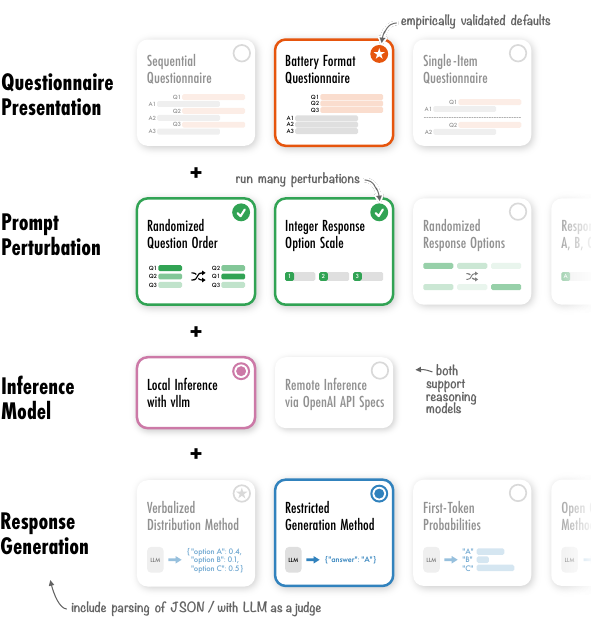}
    \caption{\textbf{\texttt{QSTN} Facilitates Easy To Customize and Robust Questionnaire Inference with LLMs.} \texttt{QSTN} provides a fully modular pipeline with different ways to present the questionnaire, prompt perturbations and to choose a response generation method, with automatic parsing. Both local and remote inference are supported.}
    \label{fig:figure_1}
    \vspace{-0.4cm}
\end{figure}



Questionnaires have become an important format to probe, assess, and utilize large language models (LLMs) via prompts. Questionnaire-like prompts have been a popular way to evaluate LLMs on tasks such as common knowledge understanding \cite{hendrycks2021measuring}, language comprehension \cite{hu2023finegrained,sravanthi2024pub,kimyeeun2024developing}, and mathematical reasoning \cite{satpute2024can,wei2023cmath}. Other work uses existing questionnaires to evaluate LLMs' values; for example, political bias \cite{rottger2024political, rozado2024political}, personality traits \cite{jiang2024personallm,shu2024dont,pellert2024ai}, or psychometric profiles \cite{ye2025large}.
With the increasing capability of LLMs, researchers have found additional use cases, such as the creation of synthetic survey responses \cite{argyle2023out, ma2024potential} or data annotation \cite{tan2024large}. 

Despite the widespread use of questionnaire-like prompts, concerns have been raised about the robustness of LLM responses to such prompts. The closed-ended responses of an LLM can vary strongly from its open-ended responses~\cite{rottger2024political, wang2024answerc}, LLM responses can be biased towards specific survey response options~\cite{tjuatja2024llms, rupprecht2025prompt}, and downstream performance is strongly affected by small changes in the questionnaire configuration~\cite{cummins2025threat, ahnert2025survey}.


To address and investigate some of these concerns, \textbf{we present \texttt{QSTN}} (pronounced ``Question'') - a Python framework designed to \textbf{facilitate the execution of questionnaire-style experiments with LLMs.} \texttt{QSTN} simplifies the process of creating robust variations of question prompts and answer generation methods, thereby facilitating reproducibility and the analysis of the reliability of LLM-based questionnaire research. \texttt{QSTN} provides a complete, modular pipeline, as depicted in Figure \ref{fig:figure_1}, for creating the questionnaire presentation, adjusting various parts of the prompt with perturbations, choosing the response generation method, performing inference, and finally, parsing the generated text. 
We evaluate our framework on more than 40 million survey responses and find that the controlled variation of the experiment pipeline can increase the alignment of generated responses with human survey answers and that the responses can be obtained for a fraction of the compute cost.

\begin{itemize}[leftmargin=15pt, itemsep=0.05em]
    \item[\small\faIcon{code}] \textbf{Python package under MIT license:}\\\href{https://github.com/dess-mannheim/QSTN}{https://github.com/dess-mannheim/QSTN}
    
    \item[\small{\faIcon[regular]{window-maximize}}] \textbf{Live GUI:} \href{https://hf.co/spaces/qstn/qstn_gui}{https://hf.co/spaces/qstn/qstn\_gui}\\or run it locally by cloning the Git repository
    
    \item[\small\faIcon{video}] \textbf{Video:} \href{https://youtu.be/uM5Q-Qmm6nQ}{https://youtu.be/uM5Q-Qmm6nQ}
\end{itemize}

\section{Core Features}


\texttt{QSTN} was developed with three objectives in mind: First, it enables \emph{robust evaluation} of and with LLMs, addressing prompt sensitivity \cite{tjuatja2024llms,dominguezolmedo2024questioning}. \texttt{QSTN} is engineered to address this challenge directly through a highly modular and configurable design. Each part of the pipeline can be exchanged independently from the other parts.

Second, \texttt{QSTN} is designed to be \emph{efficient}, so it can be used in large-scale studies. For experiments with multiple prompt variations and/or personas, we automatically utilize prefix caching and batching for local inference in vLLM \cite{Woosuk2023vllm}, and asynchronous calling with the AsyncOpenAI API \cite{openai_python_2023}.

Finally, \texttt{QSTN} is designed to be as \emph{easy to use as possible}. 
Since we maintain the common prompt format of the system prompt and user prompt, adapting a project to \texttt{QSTN} is seamless. The package offers a complete pipeline from prompt creation and inference to parsing, which can be done in only three function calls to the package. Integration with existing vLLM and OpenAI packages is straightforward.

\texttt{QSTN}'s \emph{core strength lies in its ability to systematically and easily control and vary the setup} of questionnaire-like prompting experiments. The following aspects of the experiments can be exchanged and varied by simply switching out one module for another.

\subsection{{\color{tab_orange}{$\blacksquare$}} Questionnaire Presentation}
\label{sec:questionnaire_presentation}

\begin{figure}[t]
    \centering
    \includegraphics[width=0.9\linewidth]{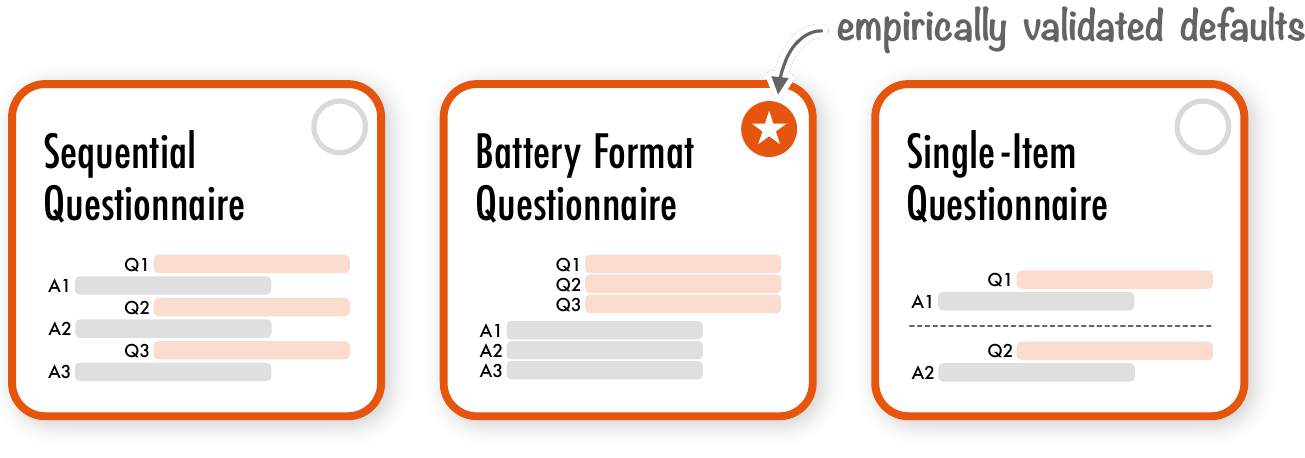}
    \caption{\textbf{\texttt{QSTN} Questionnaire Presentation Modes}}
    \vspace{-0.3cm}
    \label{fig:questionnaire_presentation}
\end{figure}

\texttt{QSTN} supports three distinct questionnaire presentation modes, as shown in Figure \ref{fig:questionnaire_presentation}: 

\begin{itemize}[leftmargin=*,noitemsep]
\item[{\color{tab_orange}{$\blacksquare$}}] \textbf{Sequential:} Each question is asked in the same conversation context in multiple, sequential chat calls.
\item[{\color{tab_orange2}{$\blacksquare$}}] \textbf{Battery:} All questions are asked in \batteryitem and the model is expected to answer all questions in one response in the same context. 
\item[{\color{tab_orange3}{$\blacksquare$}}] \textbf{Single-item:} Each question is asked in a new context, with the LLM not being aware of the previous questions and answers.

\end{itemize}

Questionnaire presentation is a fundamental decision to make when using LLMs with questionnaire-like prompts. For example, if we want to annotate data, is it better to give all annotation questions in the same prompt, or should each question be asked in a new context? There is evidence that keeping multiple tasks in the prompts can improve variety for creative writing \cite{zhang2025verbalized} and improve performance for classification tasks in moral foundations \cite{chen2025mova}. LLMs are also able to perform multiple tasks of different kinds in \batteryitem \cite{son2024multitask}, which can save computing time.


\subsection{{\color{tab_green}{$\blacksquare$}} Prompt Perturbation}
Previous studies found that LLMs synthetic survey responses are highly sensitive to prompt perturbations and exhibit biases, such as token biases, recency bias, or A-bias \cite{pezeshkpour2024large,li2025anchored,rupprecht2025prompt,dominguezolmedo2024questioning,rottger2024political}. \texttt{QSTN} can automatically randomize or reverse both the order of the questions within the survey and the order of answer options for each question to identify and mitigate these biases. This ensures that high performance is robust and independent of ordering. Previous research has found that LLMs can be sensitive to small changes in prompt format \cite{he2024does,sclar2024quantifying}. \texttt{QSTN} allows users to define custom answer label schemas (e.g., A/B/C, 1/2/3, i/ii/iii), enabling rigorous testing of a model's robustness to superficial formatting changes. QSTN can perform the following Answer Option Perturbations, which are  shown in Figure \ref{fig:prompt_perturbation}:

\begin{figure}[t]
    \centering
    \includegraphics[width=0.9\linewidth]{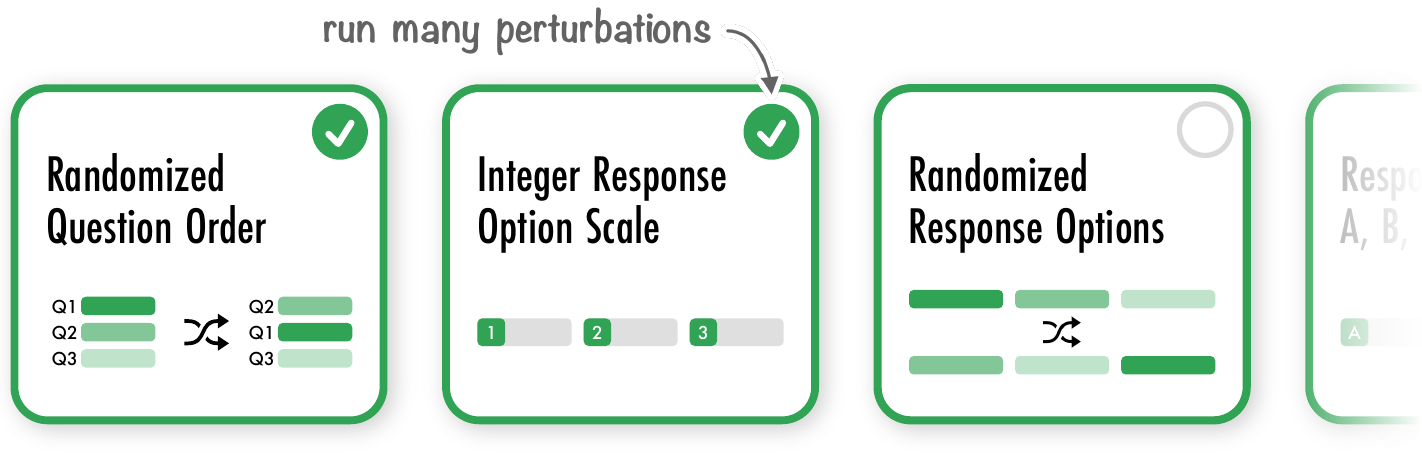}
    \caption{\textbf{\texttt{QSTN} Supported Prompt Perturbations}}
    \label{fig:prompt_perturbation}
\end{figure}

\begin{itemize}[leftmargin=*,noitemsep]
    \item[\color{tab_green}{$\blacksquare$}] \textbf{Reversed Response Order:} The order of answer options is reversed (e.g., a scale from `1: Very important' to `5: Not important' becomes `1: Not important' to `5: Very important').
    \item[\color{tab_green}{$\blacksquare$}] \textbf{Missing Refusal Option:} The ``Don't know'' or refusal option is removed from the list of choices.
    \item[\color{tab_green}{$\blacksquare$}] \textbf{Odd/Even Scale Transformation:} For scales with an even number of options, a semantically appropriate middle category is added, transforming it into an odd-numbered scale (e.g., by adding `Neutral'). Conversely, for odd-numbered scales, we remove the middle category to create an even scale and adjust the integer label accordingly.
\end{itemize}

\noindent
In addition, \texttt{QSTN} can perform the following Question Perturbations: 

\begin{itemize}[leftmargin=*,noitemsep]
    \item[\color{tab_green3}{$\blacksquare$}] \textbf{Typographical Errors:} three types of typos can be introduced: \emph{Key Typo} (replacing a character with a random one), \emph{Letter Swap} (swapping two adjacent characters in a random word), and \emph{Keyboard Typo} (replacing a character with an adjacent one on a QWERTY keyboard).
    \item[\color{tab_green3}{$\blacksquare$}] \textbf{Semantic Variations:} Additional semantic variations can be introduced while preserving the original meaning: first, by \emph{Synonym Replacement}, where a variable amount of words in the original question are replaced with synonyms. Second, through \emph{Paraphrasing} the entire question is rephrased.
\end{itemize}

\subsection{{\color{tab_blue}{$\blacksquare$}} Response Generation}

\begin{figure}[t]
    \centering
    \includegraphics[width=0.9\linewidth]{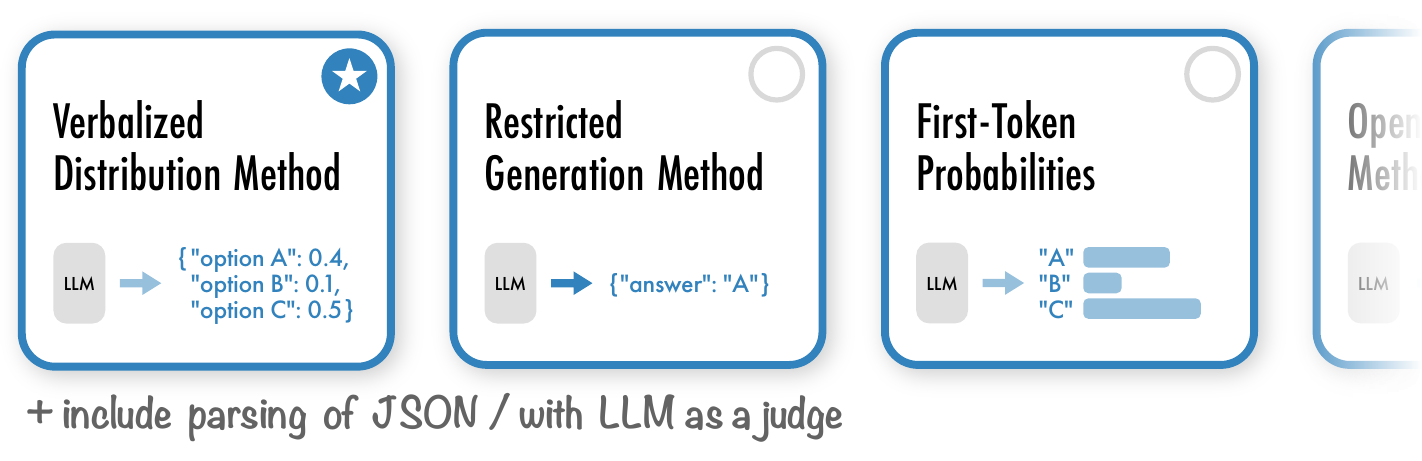}
    \caption{\textbf{\texttt{QSTN} Response Generation Methods}}
    \vspace{-0.4cm}
    \label{fig:response_generation}
\end{figure}

While generative language models are designed to generate open-ended text, previous studies have implemented various approaches to constrain LLMs to closed-ended responses~\cite[e.g.,][]{ma2024potential}.
We define \textbf{Response Generation Methods} as techniques used to elicit closed-ended responses from large language models to questionnaires~\cite{ahnert2025survey}. 
QSTN supports the following Response Generation Methods, with examples being shown in Figure \ref{fig:response_generation}:
\begin{itemize}[leftmargin=*,noitemsep]
    \item[\color{tab_blue}{$\blacksquare$}] \textbf{Token Probability-Based Methods:} Extract probabilities for response options from the output token probabilities of an LLM.
    \item[\color{tab_blue2}{$\blacksquare$}] \textbf{Restricted Generation Methods:} Force the model to respond only with designated response options using formatting instructions in the prompt and (optionally) restrict the vocabulary of the LLM through \textit{structured outputs}.
    \item[\color{tab_blue3}{$\blacksquare$}] \textbf{Open Generation Methods:} Generate open-ended responses first and then classify them in a second step.
\end{itemize}

\noindent
The Restricted Generation Methods can be used to generate exactly one of the available response options---optionally in JSON format, or with reasoning---or to generate a \textbf{verbalized distribution} of probabilities for all response options, following~\citet{meister-etal-2025-benchmarking}.
All Response Generation Methods can be adjusted to, e.g., have the model generate a prefix before token probabilities are extracted. \texttt{QSTN} includes \textbf{suitable parsers} for all generated responses: JSON \& LLM-as-a-judge.


\section{Using \texttt{QSTN}}

The package containing \texttt{QSTN} can be installed in the desired environment using pip. We support both a lightweight installation with \texttt{pip install qstn}, which only requires dependencies for API usage, and a full installation with \texttt{pip install qstn[vllm]} , which allows for local inference with vllm. \texttt{QSTN} is easily integrable into current workflows, requiring just a total of three function calls for the most basic functionality, and it still allows users to freely define their prompts. A minimum usage example is given in Listing \ref{lst:quell_example}. By simply exchanging the function in the inference step, the questionnaire presentation can be adjusted, or a different type of parser can be selected. Additionally, building on this simple example, only one more module is needed to implement controlled prompt perturbations and response generation methods.

\begin{listing}[bh!]
\begin{lstlisting}[style=mypython]
import qstn
import pandas as pd
from vllm import LLM

# 1. Prepare questionnaire and persona data
questionnaires = pd.read_csv("hf://datasets/qstn/ex/q.csv")
personas = pd.read_csv("hf://datasets/qstn/ex/p.csv")
prompt = (
    f"Please tell us how you feel about:\n"
    f"{qstn.utilities.placeholder.PROMPT_QUESTIONS}"
)
interviews = [
    qstn.prompt_builder.LLMPrompt(
        questionnaire_source=questionnaires,
        system_prompt=persona,
        prompt=prompt,
    ) for persona in personas.system_prompt]

# 2. Run Inference
model = LLM("Qwen/Qwen3-4B", max_model_len=5000)
results = qstn.survey_manager.conduct_survey_single_item(
    model, interviews, max_tokens=500
)

# 3. Parse Results
parsed_results = qstn.parser.raw_responses(results)
\end{lstlisting}

\caption{\textbf{Minimum usage example of \texttt{QSTN}.} \texttt{QSTN} can be easily integrated into existing projects, requiring just three function calls to operate. Users familiar with vllm or the OpenAI API can use the same Model/Client calls and arguments. In this example reasoning and the generated response are automatically parsed.}
\label{lst:quell_example}
\vspace{-0.4cm}
\end{listing}

\paragraph{Non-Code User Interface}

\texttt{QSTN} offers a User Interface to create and run inference with LLMs without having to program any Python code. The UI offers the same core functionality as the main framework, allowing users to upload questionnaires, systematically alter the prompt structure, set model parameters, and run inference. While the UI generally offers the same functions as the coding package, some more advanced features, such as inferencing models directly through the vllm API, are currently not supported.


\section{Evaluation}



We evaluate \texttt{QSTN} primarily on the task of generating synthetic survey responses, which is a topic of growing interest. Our results demonstrate that our proposed variations significantly influence both the alignment of synthetic data with real-world responses and computational efficiency. Across all experiments, we use the following instruction-finetuned versions of the models: LLama3 1-70B \cite{grattafiori2024llama}, Qwen3 4-30B \cite{yang2025qwen3}, Phi-4-mini \cite{abdin2024phi}, Gemma3 4-27B \cite{team2025gemma}, OLMo2 1-32B \cite{olmo20242}, Yi1.5 6B \cite{young2024yi}, and Gemini1.5 Pro \cite{team2024gemini}. We present new results and evaluations regarding questionnaire presentation and provide an overview of previous experiments that were conducted in or implemented into \texttt{QSTN}, which evaluate Prompt Perturbations and Response Generation Methods.

\subsection{{\color{tab_orange}{\textbf{\rule{1ex}{1ex}}}} Questionnaire Presentation}

We start by demonstrating that the presentation of the questionnaire significantly impacts the subpopulation alignment of generated responses with real answers. Furthermore, selecting the optimal method results in savings for both token usage and GPU time. We test three fundamentally different presentations, as described in Section \ref{sec:questionnaire_presentation}.


We base our experiments on \citet{bisbee2024synthetic}, where respondents of the ANES survey are instructed to consider a certain group and to indicate the degree to which they experience warm (positive, affectionate, etc.) or cool (negative, disdainful, etc.) feelings toward members of that group on a scale from 0 to 100. For each of the 7530 participants, we use three different seeds, which leads to a \textbf{total of 10,843,200 individual question responses} across 16 questions, 10 different models, and 3 different presentations.

We use the same prompts as in the initial study, with the addition of an instruction on how to format the output to align with the response generation method. Our full prompts can be seen in Table \ref{tab:prompt_assembly} in the Appendix.
Respondents were stratified into subpopulations based on the intersection of gender, race, and ideology (see Appendix Table~\ref{tab:subpopulation_list} for full subpopulation attributes). We measure individual alignment via Mean Absolute Error and subpopulation distributional alignment via Wasserstein distance; results are displayed in Table~\ref{tab:results_wsd_mae}. To quantify the effects of questionnaire presentation, we fitted Ordinary Least Squares and Weighted Least Squares models for MAE and Wasserstein distance, respectively. Both models include interaction terms between presentation and model, as well as fixed effects for iteration seeds. The \singleitem presentation and \texttt{Llama-3.3-70B-Instruct} serve as the reference categories.

\begin{table*}[htb]
\begin{adjustbox}{width={\textwidth}}
\begin{tabular}{lccc|ccc}
\toprule
 & \multicolumn{3}{c}{Mean Absolute Error $\downarrow$} & \multicolumn{3}{c}{Wasserstein distance $\downarrow$} \\
questionnaire presentation & {\color{tab_orange}{$\blacksquare$}} sequential & {\color{tab_orange2}{$\blacksquare$}} battery &  {\color{tab_orange3}{$\blacksquare$}} single-item & {\color{tab_orange}{$\blacksquare$}} sequential & {\color{tab_orange2}{$\blacksquare$}} battery &  {\color{tab_orange3}{$\blacksquare$}} single-item \\
\midrule
gemma-3-4b-it & 20.96 $\pm$ 0.02 & 21.92 $\pm$ 0.01 & \textbf{19.94 $\pm$ 0.02} & 16.48 $\pm$ 0.01 & 17.62 $\pm$ 0.02 & \textbf{16.39 $\pm$ 0.01} \\
gemma-3-12b-it & 18.26 $\pm$ 0.01 & \textbf{18.07 $\pm$ 0.02} & 19.11 $\pm$ 0.01 & 14.53 $\pm$ 0.02 & \textbf{13.44 $\pm$ 0.01} & 16.44 $\pm$ 0.01 \\
gemma-3-27b-it & \textbf{17.59 $\pm$ 0.01} & 17.90 $\pm$ 0.01 & 18.01 $\pm$ 0.01  & \textbf{14.00 $\pm$ 0.01} & 14.26 $\pm$ 0.01 & 15.17 $\pm$ 0.00 \\
\midrule
Llama-3.2-1B-Instruct & 30.89 $\pm$ 0.25 & \textbf{30.22 $\pm$ 0.07} & 35.69 $\pm$ 0.12 & 18.66 $\pm$ 0.33 & \textbf{18.15 $\pm$ 0.12} & 27.52 $\pm$ 0.17 \\
Llama-3.2-3B-Instruct & 24.20 $\pm$ 0.10 & \textbf{22.98 $\pm$ 0.04} & 24.32 $\pm$ 0.06  & \textbf{13.14 $\pm$ 0.11} & 13.50 $\pm$ 0.03 & 15.88 $\pm$ 0.07 \\
Llama-3.1-8B-Instruct & 21.01 $\pm$ 0.04 & 20.88 $\pm$ 0.02 & \textbf{20.87 $\pm$ 0.04} & 13.62 $\pm$ 0.02 & \textbf{12.90 $\pm$ 0.02} & 14.11 $\pm$ 0.04 \\
Llama-3.3-70B-Instruct & 18.23 $\pm$ 0.00 & \textbf{17.67 $\pm$ 0.00} & 17.87 $\pm$ 0.01 & 14.18 $\pm$ 0.00 & \textbf{13.56 $\pm$ 0.01} & 14.73 $\pm$ 0.01 \\
\midrule
Phi-4-mini-instruct & 20.98 $\pm$ 0.03 & \textbf{19.72 $\pm$ 0.01} & 21.23 $\pm$ 0.03  & \textbf{11.69 $\pm$ 0.06} & 12.21 $\pm$ 0.02 & 14.56 $\pm$ 0.05 \\
\midrule
Qwen3-4B-Instruct-2507& \textbf{19.29 $\pm$ 0.02} & 20.34 $\pm$ 0.01 & 20.05 $\pm$ 0.02 & \textbf{13.75 $\pm$ 0.02} & 15.59 $\pm$ 0.01 & 15.18 $\pm$ 0.01 \\
Qwen3-30B-A3B-Instruct-2507 & 17.68 $\pm$ 0.02 & \textbf{17.67 $\pm$ 0.01} & 18.29 $\pm$ 0.01 & 13.88 $\pm$ 0.02 & \textbf{13.68 $\pm$ 0.01} & 15.21 $\pm$ 0.02 \\
\bottomrule
\end{tabular}
\end{adjustbox}
\caption{\textbf{Individual and subpopulation alignment based on {\color{tab_orange}{\textbf{\rule{1ex}{1ex}}}} questionnaire presentation.} Mean absolute error for each individual response and weighted mean Wasserstein distance across the subpopulations. Wasserstein distance significantly improves with sequential and battery presentation for most models, compared to single-item.}
\label{tab:results_wsd_mae}
\end{table*}

We find that questionnaire presentation has a substantial impact on distributional alignment, whereas the effects on individual-level accuracy, while statistically significant, are practically marginal. For the reference model, the \batteryitem presentation yields the strongest improvement in subpopulation alignment ($\beta_{\text{WD}} = -1.17, p < 0.01$), representing an approximate $8\%$ better alignment than with the \singleitem presentation. This effect is consistent across the large models we tested, as the interaction effect for both \texttt{Qwen-30B} and \texttt{Gemma-27B} was not statistically significant. However, for smaller models, the effect is highly architecture-dependent: \texttt{Phi-4-mini} achieves the best overall alignment in our experiment using the \sequentialitem presentation, whereas \texttt{gemma-3-4b} achieves the best alignment with \singleitem presentation.


Considering the large differences in tokens and compute time between the presentation methods (shown in Table \ref{tab:token_time}), \textbf{we recommend the \batteryitem presentation as the default for future questionnaire-based experiments with large persona prompts}. However, thorough tests should be conducted to ensure that performance is comparable to other presentations for the specific model and task at hand. \texttt{QSTN} makes these validation experiments accessible by requiring just a single method change in the pipeline.


\subsection{{\color{tab_green}{\textbf{\rule{1ex}{1ex}}}} Prompt Perturbation}
In previous research \cite{rupprecht2025prompt}, we found a consistent recency bias in all nine models tested, favoring the same answer option when placed at the end of the options list instead of the beginning. This effect was substantial, with the selection frequency of the semantically same option increasing by more than 20 times for \texttt{Llama-3.1-8B} when moved to the last position, while all other configurations, such as question and prompt phrasing, were kept constant.

All models facing prompt perturbations showed some level of non-robust responses, whereas larger models such as \texttt{Llama-3.3-70B} and \texttt{Gemini-1.5-Pro} respond more robustly. The magnitude of the effect of perturbations (e.g., on the answer option or the question phrasing) on response robustness mainly depends on the type of perturbation applied. We identified that some of the {\color{tab_green}{$\blacksquare$}} Answer Option Perturbations and {\color{tab_green2}{$\blacksquare$}} Question Perturbations have a larger impact on response robustness than others (see Table~\ref{tab:perturbation_robustness}). Reversing the answer options or introducing typos or paraphrasing the questions is more harmful to robustness than swapping characters within a word or removing the refusal category. In addition, we found that 67\% and 89\% of models select the middle category significantly more often when a 5- or 11-point Likert scale is provided, respectively.

These findings underline the importance of robustness checks, e.g., through prompt perturbations. \texttt{QSTN} allows the user to apply various perturbations automatically to any questionnaire presented and thus assess the response robustness of the LLM.

\begin{table}[tb]
\centering
\begin{adjustbox}{width={\columnwidth}}
\begin{tabular}{lcccc}
\toprule
\textbf{Presentation} & \textbf{Calls} & \textbf{Input T.} & \textbf{Output T.} & \textbf{Inference Time} \\
\midrule
\sequentialitem     & 16 & 8216 & 288 & 09:29:05 \\
\batteryitem   & 1  & 723  & 142 & 01:34:45 \\
\singleitem     & 16 & 4288 & 288 & 03:22:23 \\
\bottomrule
\end{tabular}
\end{adjustbox}
\caption{\textbf{API Calls, Tokens and inference time of different {\color{tab_orange}{\textbf{\rule{1ex}{1ex}}}} questionnaire presentations.} We report the number of API calls, tokens and inference time for the largest model LLama-3.3-70B-Instruct. The tokens are calculated on one persona and the time is measured by a whole run of 7530 personas with 3 seeds. All experiments have been conducted with vllm on two 2 NVIDIA H100 GPUs (tensor-parallel).}
\label{tab:token_time}
\end{table}

\subsection{{\color{tab_blue}{\textbf{\rule{1ex}{1ex}}}} Response Generation Methods}
\label{sec:ResponseGenerationMethod}

To investigate the impact of Response Generation Methods on generated questionnaire responses, we \textbf{predict survey responses to questions of political attitudes} in the American National Election Study~\cite{anes_2016_2016}, the German Longitudinal Election Study~\cite{gles_gles_2017, gles_gles_2025}, and the American Trends Panel~\cite{atp_american_2021}. We thereby partially replicate the studies by~\citet{argyle2023out}, \citet{vonderheyde2024vox}, and \citet{santurkar2023whose}, while extending them to include additional Response Generation Methods. We compare 8 Response Generation Methods on 10 open-weight LLMs, including reasoning models. For robustness, we include 4 prompt variations, 3 random seeds for temperature-scaled decoding, as well as greedy decoding. Overall, \textbf{we simulate 32 mio. survey responses with \texttt{QSTN}}, and evaluate their alignment with human survey responses on individual and subpopulations levels. For subpopulation-level alignment, we split the set of respondents into subpopulations by considering all unique values of all persona attributes that were included in the studies we replicate, e.g., women \& men, people from different states, etc. We report the subpopulation-level alignment on categorical response distributions using total variation distance~\cite[see also][]{meister-etal-2025-benchmarking, baan-etal-2022-stop}.

\begin{table}[t]
\centering
\resizebox{\columnwidth}{!}{%
\begin{tabular}{lccc|ccccc}
\toprule
& \multicolumn{3}{c}{{\textcolor{tab_green}{\textbf{\rule{0.9ex}{0.9ex}}}}\;\textbf{Answer Options}} & \multicolumn{5}{c}{{\textcolor{tab_green2}{\textbf{\rule{0.9ex}{0.9ex}}}}\;\textbf{Question Perturbations}} \\
\cmidrule(lr){2-4} \cmidrule(lr){5-9}
\textbf{Model} & \textbf{(1)} & \textbf{(2)} & \textbf{(3)} & \textbf{(4)} & \textbf{(5)} & \textbf{(6)} & \textbf{(7)} & \textbf{(8)} \\ \midrule
Llama-3.3-70B & 0.50 & 0.73 & \textbf{0.60} & 0.52 & \textbf{0.76} & 0.58 & 0.58 & \textbf{0.66} \\
Llama-3.1-8B & 0.08 & 0.39 & 0.27 & 0.32 & 0.31 & 0.23 & 0.32 & 0.16\\
Llama-3.2-3B & 0.10 & 0.11 & 0.16 & 0.10 & 0.16 & 0.18 & 0.23 & 0.10\\
Llama-3.2-1B & 0.00 & 0.11 & 0.03 & 0.05 & 0.11 & 0.00 & 0.13 & 0.02 \\
Gemini-1.5-Pro & \textbf{0.69} & 0.76 & 0.55 & \textbf{0.68} & 0.73 & \textbf{0.66} & 0.60 & 0.55\\
Phi-3.5-mini & 0.53 & \textbf{0.81} & 0.45 & 0.50 & 0.61 & 0.47 & \textbf{0.71} & 0.53 \\
Mistral-7B-v0.3 & 0.68 & \textbf{0.81} & 0.53 & 0.58 & 0.65 & 0.60 & \textbf{0.71} & 0.53 \\
Qwen-2.5-7B & 0.32 & 0.48 & 0.45 & 0.48 & 0.65 & 0.45 & 0.55 & 0.44 \\
Yi-1.5-6B & 0.47 & 0.68 & 0.55 & 0.50 & 0.50 & 0.45 & 0.65 & 0.29 \\ \bottomrule
\end{tabular}%
}
\caption{\textbf{Impact of {\color{tab_green}$\blacksquare$}~Answer Option and {\color{tab_green2}$\blacksquare$}~Question Perturbations on the Response Robustness of different LLMs $(\uparrow)$.} Share of fully robust responses per model. Bold indicates the highest robustness score for that perturbation type. Perturbation Keys: (1) Reversed Answer Options, (2) Missing Refusal, (3) Even Scale, (4) Key Typos, (5) Letter Swap, (6) Keyboard Typos, (7) Synonyms, (8) Paraphrase}
\label{tab:perturbation_robustness}
\end{table}

Table~\ref{tab:response_ols_short} shows selected OLS regression coefficients for subpopulation-level alignment. We find that the Verbalized Distribution Method yields significant improvements on most datasets. In combination with the individual-level alignment results presented in~\citet{ahnert2025survey}, we conclude that: (i) the \textbf{choice of Survey Response Generation Method should be well-justified} for \textit{in-silico} surveys, since we find significant differences between these methods. (ii) We \textbf{do not recommend the use of Token Probability-Based Methods}, as they generate misaligned survey responses. (iii) For predicting closed-ended survey responses, we suggest to \textbf{consider Restricted Generation Methods first,} as they consistently show significant improvement over other methods while also being more computationally efficient than Open Generation Methods.

\begin{table}[t!]
    \centering
    \footnotesize
    \begin{tblr}{
      colsep=3pt,
      colspec={Q[l, colsep=6pt] X[l] X[l] X[l] X[l]}, 
      rowsep=1.5pt,
      row{1,2}={font=\bfseries, rowsep=0.5pt},
      width=\linewidth,
      hline{3,4}={black},
      hborder{1,3,4}={belowspace=3pt},
      hline{1,7}={black, 0.65pt},
    }
Response & ANES & GLES & GLES & ATP \\
Generation Method & 2016 & 2017 & 2025 & 2021 \\
Intercept & .374* & .312* & .288* & .503* \\
{\textcolor{tab_blue}{\textbf{\rule{0.9ex}{0.9ex}}}}\;First-Token Prob. & -.003 & .147* & .194* & -.049* \\
{\textcolor{tab_blue2}{\textbf{\rule{0.9ex}{0.9ex}}}}\;Verbalized Distrib. & \textbf{-.074*} & \textbf{-.057*} & -.013 & \textbf{-.168*} \\
{\textcolor{tab_blue3}{\textbf{\rule{0.9ex}{0.9ex}}}}\;Open-Ended Distrib. & -.006 & -.052* & \textbf{-.037*} & -.082* \\
    \end{tblr}
    \caption{\textbf{Impact of {\color{tab_blue}$\blacksquare$}~Response Generation Methods on Subpopulation-Level Alignment $(\downarrow)$.}
    OLS regression coefficients by dataset with total variation distance $(\downarrow)$ as the dependent variable and Survey Response Generation Method, prompt perturbation, and LLM as independent variables. We show coefficients for selected Response Generation Methods (Reference: Restricted Choice)---see Appendix~\ref{app:response_ols} for all coefficients and more details on OLS model choice. \textbf{The Verbalized Distribution Method leads to significant improvements.} $\text{*}\, p < 0.05\;\text{(Benjamini–Hochberg corrected)}$}
    \label{tab:response_ols_short}
    \vspace{-1.7pt}
\end{table}

\section{Related Work}

Due to the importance of controlled prompt perturbation, a number of frameworks have started to address this issue. In general, \texttt{QSTN} supports controlled variation and combines it with the pipeline to allow for automatic parsing of all prompt variations. Additionally, as \texttt{QSTN} allows for modular prompts, these frameworks can be used in conjunction with it. PromptSuite \cite{habba2025promptsuite} focuses on prompt perturbation through paraphrasing and formatting. PromptSource \cite{bach2022promptsource} is a framework for making and sharing different types of natural language prompts. Prompt-Agnostic Fine-Tuning (PAFT) \cite{wei2025paft} varies prompts in the fine-tuning process rather than during inference.

There are also frameworks that model the entire pipeline of LLM experiments, similar to \texttt{QSTN}. Unitxt \cite{bandel2024unitxt} is an open-source Python framework for data processing pipelines. While powerful, it requires users to understand the Unitxt operator language, which can add cognitive overhead. The EDSL framework \cite{Horton2024EDSL} can be used to run surveys with LLMs, but it does not provide full freedom over the exact system prompt or prompt and the Response Generation Method.

\section{Conclusion}

We introduce \texttt{QSTN}, a Python framework designed to make LLM inference with questionnaires more robust. Our evaluation demonstrates that by enabling controlled variations in the generation process, \texttt{QSTN} can significantly improve the alignment of generated responses with human answers while reducing inference costs. A core feature of \texttt{QSTN} is its modularity, allowing researchers to easily vary their experimental setup with only minimal additional coding effort. The framework is broadly applicable to tasks such as data annotation, synthetic data generation, persona studies, and the analysis of LLM behavior itself.



\section*{Limitations}

Currently, our evaluation is primarily focused on the creation of synthetic survey responses. We hope that by releasing \texttt{QSTN} to the open-source community, more robust experiments can be conducted in other application domains.
While we support a variety of different Response Generation Methods and parsing options, we currently do not support every type of structured output; for example, we do not support output that is guided by a regex pattern or context free grammar. As such, not every type of experiment can currently be conducted in \texttt{QSTN}. We hope that by making the project open-source, we will be able to support more ways to conduct experiments.
Additionally, while we plan to add support for non-instruct models, they are currently not supported.

\bibliography{custom}

\begin{thebibliography}{51}
\providecommand{\natexlab}[1]{#1}

\bibitem[{Abdin et~al.(2024)Abdin, Aneja, Behl, Bubeck, Eldan, Gunasekar, Harrison, Hewett, Javaheripi, Kauffmann et~al.}]{abdin2024phi}
Marah Abdin, Jyoti Aneja, Harkirat Behl, S{\'e}bastien Bubeck, Ronen Eldan, Suriya Gunasekar, Michael Harrison, Russell~J Hewett, Mojan Javaheripi, Piero Kauffmann, and 1 others. 2024.
\newblock \href {https://arxiv.org/abs/2412.08905} {Phi-4 technical report}.
\newblock \emph{arXiv preprint arXiv:2412.08905}.

\bibitem[{Ahnert et~al.(2025)Ahnert, Haensch, Plank, and Strohmaier}]{ahnert2025survey}
Georg Ahnert, Anna-Carolina Haensch, Barbara Plank, and Markus Strohmaier. 2025.
\newblock \href {https://arxiv.org/abs/2510.11586} {Survey response generation: Generating closed-ended survey responses in-silico with large language models}.
\newblock \emph{arXiv preprint arXiv:2510.11586}.

\bibitem[{{ANES}(2016)}]{anes_2016_2016}
{ANES}. 2016.
\newblock \href {https://electionstudies.org/data-center/2016-time-series-study/} {2016 {Time} {Series} {Study}}.

\bibitem[{Argyle et~al.(2023)Argyle, Busby, Fulda, Gubler, Rytting, and Wingate}]{argyle2023out}
Lisa~P. Argyle, Ethan~C. Busby, Nancy Fulda, Joshua~R. Gubler, Christopher Rytting, and David Wingate. 2023.
\newblock \href {https://doi.org/10.1017/pan.2023.2} {Out of one, many: Using language models to simulate human samples}.
\newblock \emph{Political Analysis}, 31(3):337–351.

\bibitem[{{ATP}(2021)}]{atp_american_2021}
{ATP}. 2021.
\newblock \href {https://www.pewresearch.org/the-american-trends-panel/} {The {American} {Trends} {Panel}}.

\bibitem[{Baan et~al.(2022)Baan, Aziz, Plank, and Fernandez}]{baan-etal-2022-stop}
Joris Baan, Wilker Aziz, Barbara Plank, and Raquel Fernandez. 2022.
\newblock \href {https://doi.org/10.18653/v1/2022.emnlp-main.124} {Stop measuring calibration when humans disagree}.
\newblock In \emph{Proceedings of the 2022 Conference on Empirical Methods in Natural Language Processing}, pages 1892--1915, Abu Dhabi, United Arab Emirates. Association for Computational Linguistics.

\bibitem[{Bach et~al.(2022)Bach, Sanh, Yong, Webson, Raffel, Nayak, Sharma, Kim, Bari, Fevry, Alyafeai, Dey, Santilli, Sun, Ben-David, Xu, Chhablani, Wang, Fries, Al-shaibani, Sharma, Thakker, Almubarak, Tang, Tang, Jiang, and Rush}]{bach2022promptsource}
Stephen~H. Bach, Victor Sanh, Zheng-Xin Yong, Albert Webson, Colin Raffel, Nihal~V. Nayak, Abheesht Sharma, Taewoon Kim, M~Saiful Bari, Thibault Fevry, Zaid Alyafeai, Manan Dey, Andrea Santilli, Zhiqing Sun, Srulik Ben-David, Canwen Xu, Gunjan Chhablani, Han Wang, Jason~Alan Fries, and 8 others. 2022.
\newblock \href {https://arxiv.org/abs/2202.01279} {Promptsource: An integrated development environment and repository for natural language prompts}.
\newblock \emph{Preprint}, arXiv:2202.01279.

\bibitem[{Bandel et~al.(2024)Bandel, Perlitz, Venezian, Friedman, Arviv, Orbach, Don-Yehiya, Sheinwald, Gera, Choshen, Shmueli-Scheuer, and Katz}]{bandel2024unitxt}
Elron Bandel, Yotam Perlitz, Elad Venezian, Roni Friedman, Ofir Arviv, Matan Orbach, Shachar Don-Yehiya, Dafna Sheinwald, Ariel Gera, Leshem Choshen, Michal Shmueli-Scheuer, and Yoav Katz. 2024.
\newblock \href {https://aclanthology.org/2024.naacl-demo.21} {Unitxt: Flexible, shareable and reusable data preparation and evaluation for generative {AI}}.
\newblock In \emph{Proceedings of the 2024 Conference of the North American Chapter of the Association for Computational Linguistics: Human Language Technologies (Volume 3: System Demonstrations)}, pages 207--215, Mexico City, Mexico. Association for Computational Linguistics.

\bibitem[{Bisbee et~al.(2024)Bisbee, Clinton, Dorff, Kenkel, and Larson}]{bisbee2024synthetic}
James Bisbee, Joshua~D. Clinton, Cassy Dorff, Brenton Kenkel, and Jennifer~M. Larson. 2024.
\newblock \href {https://doi.org/10.1017/pan.2024.5} {Synthetic replacements for human survey data? the perils of large language models}.
\newblock \emph{Political Analysis}, 32(4):401–416.

\bibitem[{Chen et~al.(2025)Chen, Sun, Li, Nguyen, Yao, Yi, Xie, Tan, and Xie}]{chen2025mova}
Ziyu Chen, Junfei Sun, Chenxi Li, Tuan~Dung Nguyen, Jing Yao, Xiaoyuan Yi, Xing Xie, Chenhao Tan, and Lexing Xie. 2025.
\newblock \href {https://doi.org/10.18653/v1/2025.emnlp-main.1687} {{M}o{V}a: Towards generalizable classification of human morals and values}.
\newblock In \emph{Proceedings of the 2025 Conference on Empirical Methods in Natural Language Processing}, pages 33204--33248, Suzhou, China. Association for Computational Linguistics.

\bibitem[{Cummins(2025)}]{cummins2025threat}
Jamie Cummins. 2025.
\newblock \href {https://arxiv.org/abs/2509.13397} {The threat of analytic flexibility in using large language models to simulate human data: A call to attention}.
\newblock \emph{arXiv preprint arXiv:2509.13397}.

\bibitem[{Dominguez-Olmedo et~al.(2024)Dominguez-Olmedo, Hardt, and Mendler-D\"{u}nner}]{dominguezolmedo2024questioning}
Ricardo Dominguez-Olmedo, Moritz Hardt, and Celestine Mendler-D\"{u}nner. 2024.
\newblock \href {https://doi.org/10.52202/079017-1458} {Questioning the survey responses of large language models}.
\newblock In \emph{Advances in Neural Information Processing Systems}, volume~37, pages 45850--45878. Curran Associates, Inc.

\bibitem[{{GLES}(2017)}]{gles_gles_2017}
{GLES}. 2017.
\newblock \href {https://doi.org/10.4232/1.13235} {{GLES} 2017 {Post}-{Election} {Cross} {Section}}.

\bibitem[{{GLES}(2025)}]{gles_gles_2025}
{GLES}. 2025.
\newblock \href {https://doi.org/10.4232/5.ZA10100.1.0.0} {{GLES} 2025 {Post}-{Election} {Cross} {Section}}.

\bibitem[{Grattafiori et~al.(2024)Grattafiori, Dubey, Jauhri, Pandey, Kadian, Al-Dahle, Letman, Mathur, Schelten, Vaughan et~al.}]{grattafiori2024llama}
Aaron Grattafiori, Abhimanyu Dubey, Abhinav Jauhri, Abhinav Pandey, Abhishek Kadian, Ahmad Al-Dahle, Aiesha Letman, Akhil Mathur, Alan Schelten, Alex Vaughan, and 1 others. 2024.
\newblock \href {https://arxiv.org/abs/2407.21783} {The llama 3 herd of models}.
\newblock \emph{arXiv preprint arXiv:2407.21783}.

\bibitem[{Habba et~al.(2025)Habba, Dahan, Lior, and Stanovsky}]{habba2025promptsuite}
Eliya Habba, Noam Dahan, Gili Lior, and Gabriel Stanovsky. 2025.
\newblock \href {https://doi.org/10.18653/v1/2025.emnlp-demos.19} {{P}rompt{S}uite: A task-agnostic framework for multi-prompt generation}.
\newblock In \emph{Proceedings of the 2025 Conference on Empirical Methods in Natural Language Processing: System Demonstrations}, pages 254--263, Suzhou, China. Association for Computational Linguistics.

\bibitem[{He et~al.(2024)He, Rungta, Koleczek, Sekhon, Wang, and Hasan}]{he2024does}
Jia He, Mukund Rungta, David Koleczek, Arshdeep Sekhon, Franklin~X Wang, and Sadid Hasan. 2024.
\newblock \href {https://arxiv.org/abs/2411.10541} {Does prompt formatting have any impact on llm performance?}
\newblock \emph{arXiv preprint arXiv:2411.10541}.

\bibitem[{Hendrycks et~al.(2021)Hendrycks, Burns, Basart, Zou, Mazeika, Song, and Steinhardt}]{hendrycks2021measuring}
Dan Hendrycks, Collin Burns, Steven Basart, Andy Zou, Mantas Mazeika, Dawn Song, and Jacob Steinhardt. 2021.
\newblock \href {https://openreview.net/forum?id=d7KBjmI3GmQ} {Measuring massive multitask language understanding}.
\newblock In \emph{International Conference on Learning Representations}.

\bibitem[{Horton and Horton(2024)}]{Horton2024EDSL}
John Horton and Robin Horton. 2024.
\newblock \href {https://github.com/expectedparrot/edsl} {Edsl: Expected parrot domain specific language for ai powered social science}.
\newblock Whitepaper, Expected Parrot.

\bibitem[{Hu et~al.(2023)Hu, Floyd, Jouravlev, Fedorenko, and Gibson}]{hu2023finegrained}
Jennifer Hu, Sammy Floyd, Olessia Jouravlev, Evelina Fedorenko, and Edward Gibson. 2023.
\newblock \href {https://doi.org/10.18653/v1/2023.acl-long.230} {A fine-grained comparison of pragmatic language understanding in humans and language models}.
\newblock In \emph{Proceedings of the 61st Annual Meeting of the Association for Computational Linguistics (Volume 1: Long Papers)}, pages 4194--4213, Toronto, Canada. Association for Computational Linguistics.

\bibitem[{Jiang et~al.(2024)Jiang, Zhang, Cao, Breazeal, Roy, and Kabbara}]{jiang2024personallm}
Hang Jiang, Xiajie Zhang, Xubo Cao, Cynthia Breazeal, Deb Roy, and Jad Kabbara. 2024.
\newblock \href {https://doi.org/10.18653/v1/2024.findings-naacl.229} {{P}ersona{LLM}: Investigating the ability of large language models to express personality traits}.
\newblock In \emph{Findings of the Association for Computational Linguistics: NAACL 2024}, pages 3605--3627, Mexico City, Mexico. Association for Computational Linguistics.

\bibitem[{Kim et~al.(2024)Kim, Choi, Choi, Choi, Park, and Hwang}]{kimyeeun2024developing}
Yeeun Kim, Youngrok Choi, Eunkyung Choi, JinHwan Choi, Hai~Jin Park, and Wonseok Hwang. 2024.
\newblock \href {https://doi.org/10.18653/v1/2024.findings-emnlp.319} {Developing a pragmatic benchmark for assessing {K}orean legal language understanding in large language models}.
\newblock In \emph{Findings of the Association for Computational Linguistics: EMNLP 2024}, pages 5573--5595, Miami, Florida, USA. Association for Computational Linguistics.

\bibitem[{Kwon et~al.(2023)Kwon, Li, Zhuang, Sheng, Zheng, Yu, Gonzalez, Zhang, and Stoica}]{Woosuk2023vllm}
Woosuk Kwon, Zhuohan Li, Siyuan Zhuang, Ying Sheng, Lianmin Zheng, Cody~Hao Yu, Joseph Gonzalez, Hao Zhang, and Ion Stoica. 2023.
\newblock \href {https://doi.org/10.1145/3600006.3613165} {Efficient memory management for large language model serving with pagedattention}.
\newblock In \emph{Proceedings of the 29th Symposium on Operating Systems Principles}, SOSP '23, page 611–626, New York, NY, USA. Association for Computing Machinery.

\bibitem[{Li and Gao(2025)}]{li2025anchored}
Ruizhe Li and Yanjun Gao. 2025.
\newblock \href {https://doi.org/10.18653/v1/2025.findings-acl.124} {Anchored answers: Unravelling positional bias in {GPT}-2{'}s multiple-choice questions}.
\newblock In \emph{Findings of the Association for Computational Linguistics: ACL 2025}, pages 2439--2465, Vienna, Austria. Association for Computational Linguistics.

\bibitem[{Ma et~al.(2024)Ma, Wang, Hu, Haensch, Hedderich, Plank, and Kreuter}]{ma2024potential}
Bolei Ma, Xinpeng Wang, Tiancheng Hu, Anna-Carolina Haensch, Michael Hedderich, Barbara Plank, and Frauke Kreuter. 2024.
\newblock \href {https://aclanthology.org/2024.findings-emnlp.513/} {The potential and challenges of evaluating attitudes, opinions, and values in large language models}.
\newblock In \emph{Findings of the Association for Computational Linguistics: EMNLP 2024}, pages 8783--8805.

\bibitem[{Meister et~al.(2025)Meister, Guestrin, and Hashimoto}]{meister-etal-2025-benchmarking}
Nicole Meister, Carlos Guestrin, and Tatsunori Hashimoto. 2025.
\newblock \href {https://doi.org/10.18653/v1/2025.naacl-long.2} {Benchmarking distributional alignment of large language models}.
\newblock In \emph{Proceedings of the 2025 Conference of the Nations of the Americas Chapter of the Association for Computational Linguistics: Human Language Technologies (Volume 1: Long Papers)}, pages 24--49, Albuquerque, New Mexico. Association for Computational Linguistics.

\bibitem[{OLMo et~al.(2024)OLMo, Walsh, Soldaini, Groeneveld, Lo, Arora, Bhagia, Gu, Huang, Jordan et~al.}]{olmo20242}
Team OLMo, Pete Walsh, Luca Soldaini, Dirk Groeneveld, Kyle Lo, Shane Arora, Akshita Bhagia, Yuling Gu, Shengyi Huang, Matt Jordan, and 1 others. 2024.
\newblock \href {https://arxiv.org/abs/2501.00656} {2 olmo 2 furious}.
\newblock \emph{arXiv preprint arXiv:2501.00656}.

\bibitem[{OpenAI(2023)}]{openai_python_2023}
OpenAI. 2023.
\newblock \href {https://github.com/openai/openai-python} {{OpenAI Python Library}}.

\bibitem[{Pellert et~al.(2024)Pellert, Lechner, Wagner, Rammstedt, and Strohmaier}]{pellert2024ai}
Max Pellert, Clemens~M. Lechner, Claudia Wagner, Beatrice Rammstedt, and Markus Strohmaier. 2024.
\newblock \href {https://doi.org/10.1177/17456916231214460} {Ai psychometrics: Assessing the psychological profiles of large language models through psychometric inventories}.
\newblock \emph{Perspectives on Psychological Science}, 19(5):808--826.
\newblock Epub 2024 Jan 2.

\bibitem[{Pezeshkpour and Hruschka(2024)}]{pezeshkpour2024large}
Pouya Pezeshkpour and Estevam Hruschka. 2024.
\newblock \href {https://doi.org/10.18653/v1/2024.findings-naacl.130} {Large language models sensitivity to the order of options in multiple-choice questions}.
\newblock In \emph{Findings of the Association for Computational Linguistics: NAACL 2024}, pages 2006--2017, Mexico City, Mexico. Association for Computational Linguistics.

\bibitem[{R{\"o}ttger et~al.(2024)R{\"o}ttger, Hofmann, Pyatkin, Hinck, Kirk, Schuetze, and Hovy}]{rottger2024political}
Paul R{\"o}ttger, Valentin Hofmann, Valentina Pyatkin, Musashi Hinck, Hannah Kirk, Hinrich Schuetze, and Dirk Hovy. 2024.
\newblock \href {https://doi.org/10.18653/v1/2024.acl-long.816} {Political compass or spinning arrow? towards more meaningful evaluations for values and opinions in large language models}.
\newblock In \emph{Proceedings of the 62nd Annual Meeting of the Association for Computational Linguistics (Volume 1: Long Papers)}, pages 15295--15311, Bangkok, Thailand. Association for Computational Linguistics.

\bibitem[{Rozado(2024)}]{rozado2024political}
David Rozado. 2024.
\newblock \href {https://arxiv.org/abs/2402.01789} {The political preferences of llms}.
\newblock \emph{Preprint}, arXiv:2402.01789.

\bibitem[{Rupprecht et~al.(2025)Rupprecht, Ahnert, and Strohmaier}]{rupprecht2025prompt}
Jens Rupprecht, Georg Ahnert, and Markus Strohmaier. 2025.
\newblock \href {https://arxiv.org/abs/2507.07188} {Prompt perturbations reveal human-like biases in llm survey responses}.
\newblock \emph{arXiv preprint arXiv:2507.07188}.

\bibitem[{Santurkar et~al.(2023)Santurkar, Durmus, Ladhak, Lee, Liang, and Hashimoto}]{santurkar2023whose}
Shibani Santurkar, Esin Durmus, Faisal Ladhak, Cinoo Lee, Percy Liang, and Tatsunori Hashimoto. 2023.
\newblock \href {https://proceedings.mlr.press/v202/santurkar23a.html} {Whose opinions do language models reflect?}
\newblock In \emph{Proceedings of the 40th International Conference on Machine Learning}, ICML'23. JMLR.org.

\bibitem[{Satpute et~al.(2024)Satpute, Gie{\ss}ing, Greiner-Petter, Schubotz, Teschke, Aizawa, and Gipp}]{satpute2024can}
Ankit Satpute, Noah Gie{\ss}ing, Andr{\'e} Greiner-Petter, Moritz Schubotz, Olaf Teschke, Akiko Aizawa, and Bela Gipp. 2024.
\newblock \href {https://doi.org/10.1145/3626772.3657945} {Can llms master math? investigating large language models on math stack exchange}.
\newblock In \emph{Proceedings of the 47th international ACM SIGIR conference on research and development in information retrieval}, pages 2316--2320.

\bibitem[{Sclar et~al.(2024)Sclar, Choi, Tsvetkov, and Suhr}]{sclar2024quantifying}
Melanie Sclar, Yejin Choi, Yulia Tsvetkov, and Alane Suhr. 2024.
\newblock \href {https://openreview.net/forum?id=RIu5lyNXjT} {Quantifying language models' sensitivity to spurious features in prompt design or: How i learned to start worrying about prompt formatting}.
\newblock In \emph{The Twelfth International Conference on Learning Representations}.

\bibitem[{Shu et~al.(2024)Shu, Zhang, Choi, Dunagan, Logeswaran, Lee, Card, and Jurgens}]{shu2024dont}
Bangzhao Shu, Lechen Zhang, Minje Choi, Lavinia Dunagan, Lajanugen Logeswaran, Moontae Lee, Dallas Card, and David Jurgens. 2024.
\newblock \href {https://doi.org/10.18653/v1/2024.naacl-long.295} {You don`t need a personality test to know these models are unreliable: Assessing the reliability of large language models on psychometric instruments}.
\newblock In \emph{Proceedings of the 2024 Conference of the North American Chapter of the Association for Computational Linguistics: Human Language Technologies (Volume 1: Long Papers)}, pages 5263--5281, Mexico City, Mexico. Association for Computational Linguistics.

\bibitem[{Son et~al.(2024)Son, Baek, Nam, Jeong, and Kim}]{son2024multitask}
Guijin Son, SangWon Baek, Sangdae Nam, Ilgyun Jeong, and Seungone Kim. 2024.
\newblock \href {https://doi.org/10.18653/v1/2024.acl-long.304} {Multi-task inference: Can large language models follow multiple instructions at once?}
\newblock In \emph{Proceedings of the 62nd Annual Meeting of the Association for Computational Linguistics (Volume 1: Long Papers)}, pages 5606--5627, Bangkok, Thailand. Association for Computational Linguistics.

\bibitem[{Sravanthi et~al.(2024)Sravanthi, Doshi, Tankala, Murthy, Dabre, and Bhattacharyya}]{sravanthi2024pub}
Settaluri Sravanthi, Meet Doshi, Pavan Tankala, Rudra Murthy, Raj Dabre, and Pushpak Bhattacharyya. 2024.
\newblock \href {https://aclanthology.org/2024.findings-acl.719} {{PUB}: A pragmatics understanding benchmark for assessing {LLM}s{'} pragmatics capabilities}.
\newblock In \emph{Findings of the Association for Computational Linguistics ACL 2024}, pages 12075--12097, Bangkok, Thailand and virtual meeting. Association for Computational Linguistics.

\bibitem[{Tan et~al.(2024)Tan, Li, Wang, Beigi, Jiang, Bhattacharjee, Karami, Li, Cheng, and Liu}]{tan2024large}
Zhen Tan, Dawei Li, Song Wang, Alimohammad Beigi, Bohan Jiang, Amrita Bhattacharjee, Mansooreh Karami, Jundong Li, Lu~Cheng, and Huan Liu. 2024.
\newblock \href {https://doi.org/10.18653/v1/2024.emnlp-main.54} {Large language models for data annotation and synthesis: A survey}.
\newblock In \emph{Proceedings of the 2024 Conference on Empirical Methods in Natural Language Processing}, pages 930--957, Miami, Florida, USA. Association for Computational Linguistics.

\bibitem[{Team et~al.(2024)Team, Georgiev, Lei, Burnell, Bai, Gulati, Tanzer, Vincent, Pan, Wang et~al.}]{team2024gemini}
Gemini Team, Petko Georgiev, Ving~Ian Lei, Ryan Burnell, Libin Bai, Anmol Gulati, Garrett Tanzer, Damien Vincent, Zhufeng Pan, Shibo Wang, and 1 others. 2024.
\newblock \href {https://arxiv.org/abs/2403.05530} {Gemini 1.5: Unlocking multimodal understanding across millions of tokens of context}.
\newblock \emph{arXiv preprint arXiv:2403.05530}.

\bibitem[{Team et~al.(2025)Team, Kamath, Ferret, Pathak, Vieillard, Merhej, Perrin, Matejovicova, Ram{\'e}, Rivi{\`e}re et~al.}]{team2025gemma}
Gemma Team, Aishwarya Kamath, Johan Ferret, Shreya Pathak, Nino Vieillard, Ramona Merhej, Sarah Perrin, Tatiana Matejovicova, Alexandre Ram{\'e}, Morgane Rivi{\`e}re, and 1 others. 2025.
\newblock \href {https://arxiv.org/abs/2503.19786} {Gemma 3 technical report}.
\newblock \emph{arXiv preprint arXiv:2503.19786}.

\bibitem[{Tjuatja et~al.(2024)Tjuatja, Chen, Wu, Talwalkwar, and Neubig}]{tjuatja2024llms}
Lindia Tjuatja, Valerie Chen, Tongshuang Wu, Ameet Talwalkwar, and Graham Neubig. 2024.
\newblock \href {https://doi.org/10.1162/tacl_a_00685} {Do {LLM}s exhibit human-like response biases? a case study in survey design}.
\newblock \emph{Transactions of the Association for Computational Linguistics}, 12:1011--1026.

\bibitem[{von~der Heyde et~al.(2025)von~der Heyde, Haensch, and Wenz}]{vonderheyde2024vox}
Leah von~der Heyde, Anna-Carolina Haensch, and Alexander Wenz. 2025.
\newblock \href {https://doi.org/10.1177/08944393251337014} {Vox populi, vox ai? using large language models to estimate german vote choice}.
\newblock \emph{Social Science Computer Review}, 0(0):1--23.

\bibitem[{Wang et~al.(2024)Wang, Ma, Hu, Weber-Genzel, R{\"o}ttger, Kreuter, Hovy, and Plank}]{wang2024answerc}
Xinpeng Wang, Bolei Ma, Chengzhi Hu, Leon Weber-Genzel, Paul R{\"o}ttger, Frauke Kreuter, Dirk Hovy, and Barbara Plank. 2024.
\newblock \href {https://doi.org/10.18653/v1/2024.findings-acl.441} {{\textquotedblleft}my answer is {C}{\textquotedblright}: First-token probabilities do not match text answers in instruction-tuned language models}.
\newblock In \emph{Findings of the Association for Computational Linguistics: ACL 2024}, pages 7407--7416, Bangkok, Thailand. Association for Computational Linguistics.

\bibitem[{Wei et~al.(2025)Wei, Ou, He, Shu, and Yu}]{wei2025paft}
Chenxing Wei, Mingwen Ou, Ying He, Yao Shu, and Fei Yu. 2025.
\newblock \href {https://doi.org/10.18653/v1/2025.emnlp-main.37} {{PAFT}: Prompt-agnostic fine-tuning}.
\newblock In \emph{Proceedings of the 2025 Conference on Empirical Methods in Natural Language Processing}, pages 694--717, Suzhou, China. Association for Computational Linguistics.

\bibitem[{Wei et~al.(2023)Wei, Luan, Liu, Dong, and Wang}]{wei2023cmath}
Tianwen Wei, Jian Luan, Wei Liu, Shuang Dong, and Bin Wang. 2023.
\newblock \href {https://doi.org/10.48550/arXiv.2306.16636} {Cmath: Can your language model pass chinese elementary school math test?}
\newblock \emph{arXiv preprint arXiv:2306.16636}.

\bibitem[{Yang et~al.(2025)Yang, Li, Yang, Zhang, Hui, Zheng, Yu, Gao, Huang, Lv et~al.}]{yang2025qwen3}
An~Yang, Anfeng Li, Baosong Yang, Beichen Zhang, Binyuan Hui, Bo~Zheng, Bowen Yu, Chang Gao, Chengen Huang, Chenxu Lv, and 1 others. 2025.
\newblock \href {https://arxiv.org/abs/2505.09388} {Qwen3 technical report}.
\newblock \emph{arXiv preprint arXiv:2505.09388}.

\bibitem[{Ye et~al.(2025)Ye, Jin, Xie, Zhang, and Song}]{ye2025large}
Haoran Ye, Jing Jin, Yuhang Xie, Xin Zhang, and Guojie Song. 2025.
\newblock \href {https://arxiv.org/abs/2505.08245} {Large language model psychometrics: A systematic review of evaluation, validation, and enhancement}.
\newblock \emph{Preprint}, arXiv:2505.08245.

\bibitem[{Young et~al.(2024)Young, Chen, Li, Huang, Zhang, Zhang, Wang, Li, Zhu, Chen et~al.}]{young2024yi}
Alex Young, Bei Chen, Chao Li, Chengen Huang, Ge~Zhang, Guanwei Zhang, Guoyin Wang, Heng Li, Jiangcheng Zhu, Jianqun Chen, and 1 others. 2024.
\newblock \href {https://arxiv.org/abs/2403.04652} {Yi: Open foundation models by 01. ai}.
\newblock \emph{arXiv preprint arXiv:2403.04652}.

\bibitem[{Zhang et~al.(2025)Zhang, Yu, Chong, Sicilia, Tomz, Manning, and Shi}]{zhang2025verbalized}
Jiayi Zhang, Simon Yu, Derek Chong, Anthony Sicilia, Michael~R Tomz, Christopher~D Manning, and Weiyan Shi. 2025.
\newblock \href {https://arxiv.org/abs/2510.01171} {Verbalized sampling: How to mitigate mode collapse and unlock llm diversity}.
\newblock \emph{arXiv preprint arXiv:2510.01171}.

\end{thebibliography}

\appendix

\section{Questionnaire Presentation}
\label{app:question}

As another measure of individual evaluation of predictions, we calculate the Pearson correlation between predictions and the ground truth and present it in Table \ref{tab:correlation_stats}. Similarly to the Mean Absolute Error, we see little difference in the performance of the different questionnaire presentations.

\begin{table}[htb]
\begin{adjustbox}{width={\columnwidth}}
\begin{tabular}{lccc}
\toprule
questionnaire presentation & {\color{tab_orange}{$\blacksquare$}} sequential & {\color{tab_orange2}{$\blacksquare$}} battery &  {\color{tab_orange3}{$\blacksquare$}} single-item \\
\midrule
gemma-3-4b-it & \textbf{0.59 ± 0.00} & 0.55 ± 0.00 & 0.57 ± 0.00 \\
gemma-3-12b-it & 0.62 ± 0.00 & 0.62 ± 0.00 & \textbf{0.64 ± 0.00} \\
gemma-3-27b-it & \textbf{0.62 ± 0.00} & 0.61 ± 0.00 & 0.61 ± 0.00 \\
\midrule
Llama-3.2-1B-Instruct & \textbf{0.25 ± 0.01} & 0.18 ± 0.00 & 0.10 ± 0.00 \\
Llama-3.2-3B-Instruct & 0.51 ± 0.00 & 0.49 ± 0.00 & \textbf{0.52 ± 0.00} \\
Llama-3.1-8B-Instruct & 0.56 ± 0.00 & \textbf{0.57 ± 0.00} & 0.56 ± 0.00 \\
Llama-3.3-70B-Instruct & \textbf{0.64 ± 0.00} & \textbf{0.64 ± 0.00} & \textbf{0.64 ± 0.00} \\
\midrule
Phi-4-mini-instruct & 0.48 ± 0.00 & 0.49 ± 0.00 & \textbf{0.52 ± 0.00} \\
\midrule
Qwen3-4B-Instruct-2507 & \textbf{0.60 ± 0.00} & 0.55 ± 0.00 & 0.59 ± 0.00 \\
Qwen3-30B-A3B-Instruct-2507 & \textbf{0.62 ± 0.00} & \textbf{0.62 ± 0.00} & 0.59 ± 0.00 \\
\bottomrule
\end{tabular}
\end{adjustbox}
\caption{\textbf{Mean and Standard Deviation of Pearson Correlation between Prediction and Ground Truth.} Similar to Mean Absolute Error, individual alignment measured with Pearson Correlation shows little difference between different questionnaire presentations.}
\label{tab:correlation_stats}
\end{table}

We show all attributes we considered for the subpopulation analysis for Wasserstein distance in Table \ref{tab:subpopulation_list}. The full regression results for both MAE and Wasserstein Distance can be seen in \ref{tab:regression_results_questionnaire_pres}. We report the coefficients and the Benjamini-Hochberg corrected p-values. Additionally, we want to determine if the questionnaire presentation has different effects on different questions. For this, we fit an additional Weighted Least Squares regression on all subpopulations based on the full interaction between the questionnaire presentation, the model, and the specific interview question. We set \singleitem, the biggest model Llama-3.3-70B-Instruct and the first question as the reference categories, as for this question the LLM has no answers for the other questions in context regardless of the questionnaire presentation. 

\begin{figure}[htb]
    \centering
    \includegraphics[width=\linewidth]{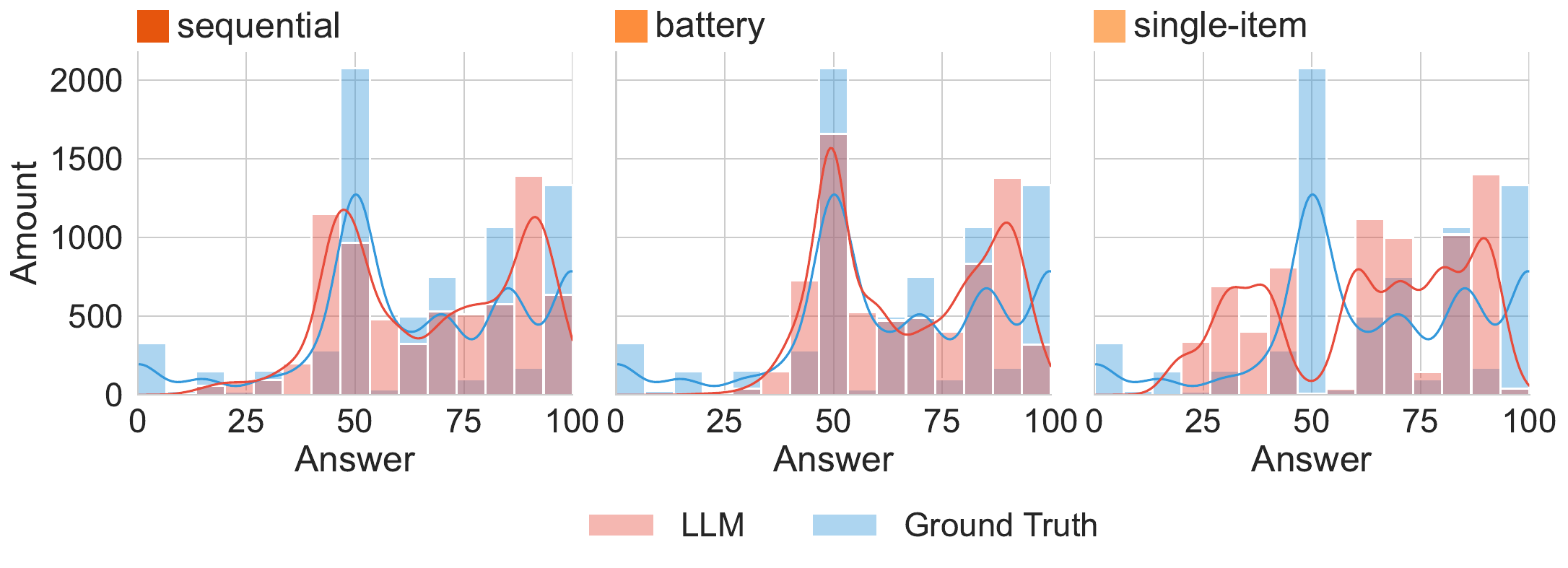}
    \caption{\textbf{Answer Distributions.} Predictions and Ground Truth Distributions across the whole population compared for LLama-3.3-70B-Instruct and the question ``How do you feel towards Gays and Lesbians?''. We can see a clear shift towards the middle for this question, when models are given context of the previous questions and answers, which aligns more closely with human answers.}
    \label{fig:distributions}
\end{figure}

All questions show improvements, and a subset of five questions shows statistically significant improvement $(p<0.05)$ when using \batteryitem presentation instead of \singleitem presentation. The largest improvement is in the question about feelings towards the group of Gays and Lesbians $(\beta=-3.82,p<0.01)$ when using \batteryitem presentation. Figure \ref{fig:distributions} visually confirms this: when previous questions and answers are included in the context, the model's response distribution aligns much more closely with the ground truth, exhibiting a similar tendency toward neutral answers. The other significant questions concern the groups of White Americans, Asian Americans, Christians, and Liberals.

\begin{table}[htb]
    \centering

    \begin{tabular}{ll}
        \toprule
        \textbf{Attribute} & \textbf{Category} \\
        \midrule
        \textbf{Race} & Non-Hispanic White \\
                      & Non-Hispanic Black \\
                      & Hispanic \\
        \addlinespace
        \textbf{Gender} & Male \\
                        & Female \\
        \addlinespace
        \textbf{Ideology} & Extremely Liberal \\
                          & Liberal \\
                          & Slightly Liberal \\
                          & Moderate \\
                          & Slightly Conservative \\
                          & Conservative \\
                          & Extremely Conservative \\
        \bottomrule
    \end{tabular}

        \caption{\textbf{Subpopulations:} We consider these subpopulations for analysis. We have the same subpopulations as the initial study by \citet{bisbee2024synthetic}.}
    \label{tab:subpopulation_list}
\end{table}

We use the same prompt as that used in \citet{bisbee2024synthetic}, as displayed in Table \ref{tab:prompt_assembly}. We adjust the output instructions to fit our choice response generation method and add all questions as instructions in the \batteryitem presentation. For all models we use the default hyperparameter settings.

\begin{table*}[htb]

    \scriptsize
    \centering
    
    \begin{adjustbox}{width={\textwidth}}
        
    \begin{tabular}{|l|p{0.5\textwidth}|}
        \hline
        \textbf{Component} & \textbf{Content} \\
        \hline
        \hline
        
        \textbf{System Prompt} 
        & "It is \texttt{\{year\}}. You are a \texttt{\{age\}} year-old, \texttt{\{marst\}}, \texttt{\{raceth\}} \texttt{\{gender\}} with \texttt{\{education\}} making \texttt{\{income\}} per year, living in the United States. You are \texttt{\{ideo\}}, \texttt{\{regis\}} \texttt{\{PID\}} who \texttt{\{interest\}} pays attention to what’s going on in government and politics." \\
        \hline
        
        \textbf{User Prompt} 
        & ``Provide responses from this person’s perspective. Use only knowledge about politics that they would have. The following questions ask about individuals’ feelings toward different groups. Responses should be given on a scale from 0 (meaning cold feelings) to 100 (meaning warm feelings). Ratings between 50 degrees and 100 degrees mean that you feel favorable and warm toward the group. Ratings between 0 degrees and 50 degrees mean that you don’t feel favorable toward the group and that you don’t care too much for that group. You would rate the group at the 50 degree mark if you don’t feel particularly warm or cold toward the group.
        
        \texttt{Output Instructions}

        \texttt{Questions}" \\
        \hline
        
        \textbf{Output Instruction}
        & You only respond in the following JSON format: 
        \begin{verbnobox}
```json
{
  "temperature": <temperature>
}
```
\end{verbnobox} 
        
        \\
        \hline

        \textbf{Output Instruction} 
        (\batteryitem)
        & You only respond in the following JSON format:
\begin{verbnobox}
```json
{
  "temperature_The Democratic Party?": <temperature_The Democratic Party?>,
  "temperature_The Republican Party?": <temperature_The Republican Party?>,
  "temperature_Democrats?": <temperature_Democrats?>,
  "temperature_Republicans?": <temperature_Republicans?>,
  "temperature_Black Americans?": <temperature_Black Americans?>,
  "temperature_White Americans?": <temperature_White Americans?>,
  "temperature_Hispanic Americans?": <temperature_Hispanic Americans?>,
  "temperature_Asian Americans?": <temperature_Asian Americans?>,
  "temperature_Muslims?": <temperature_Muslims?>,
  "temperature_Christians?": <temperature_Christians?>,
  "temperature_Immigrants?": <temperature_Immigrants?>,
  "temperature_Gays and Lesbians?": <temperature_Gays and Lesbians?>,
  "temperature_Jews?": <temperature_Jews?>,
  "temperature_Liberals?": <temperature_Liberals?>,
  "temperature_Conservatives?": <temperature_Conservatives?>,
  "temperature_Women?": <temperature_Women?>
}
```
\end{verbnobox} 
\\
        \hline
        
        \textbf{Question}
        & How do you feel towards the Republican Party? \\
        \hline
        
    \end{tabular}

    \end{adjustbox}
    \caption{\textbf{Prompt.} We use the same prompts for \sequentialitem and \singleitem and a slightly modified output instruction for the \batteryitem presentation. For \batteryitem presentation we ask all questions separated by new lines.}
    \label{tab:prompt_assembly}
\end{table*}

\begin{table*}[p]
    \centering
    \small
    \begin{tblr}{
      colsep=6pt,
      colspec={Q[l,m,3.2cm] Q[l,m] c c}, 
      rowsep=1.5pt,
      width=0.85\textwidth,
      column{1}={font=\bfseries},
      row{1}={font=\bfseries},
      row{2}={font=\bfseries, halign=c},
      hline{3,5,14,32}={-}{}, 
      hline{1,38}={-}{0.8pt}, 
      hline{2}={3-4}{0.5pt},  
      hborder{1,3,6,16,19,38}={belowspace=3pt}
    }
     &  & (1) & (2) \\
     &  & MAE (OLS) & WD Score (WLS) \\
    
    \SetCell[r=2]{h} {Questionnaire\\Presentation} 
     & \sequentialitem & 0.362** & -0.546* \\
     & \batteryitem & -0.199** & -1.166** \\

    \SetCell[r=9]{h} {Model} 
     & Llama 3.1 8B & 2.999** & -0.617* \\
     & Llama 3.2 1B & 17.822** & 12.796** \\
     & Llama 3.2 3B & 6.450** & 1.152** \\
     & Phi-4 Mini & 3.366** & -0.163 \\
     & Qwen3 30B & 0.420** & 0.488 \\
     & Qwen3 4B & 2.187** & 0.454 \\
     & Gemma 3 12B & 1.245** & 1.713** \\
     & Gemma 3 27B & 0.142** & 0.448 \\ 
     & Gemma 3 4B & 2.074** & 1.662** \\     

    \SetCell[r=18]{h} {Interactions\\(Presentation $\times$ Model)} 
     & \sequentialitem $\times$ Llama 3.1 8B & -0.216** & 0.060 \\ 
     & \batteryitem $\times$ Llama 3.1 8B & 0.213** & -0.044 \\ 
     & \sequentialitem $\times$ Llama 3.2 1B & -5.162** & -8.311** \\
     & \batteryitem $\times$ Llama 3.2 1B & -5.273** & -8.203** \\
     & \sequentialitem $\times$ Llama 3.2 3B & -0.476** & -2.195** \\
     & \batteryitem $\times$ Llama 3.2 3B & -1.136** & -1.211** \\
     & \sequentialitem $\times$ Phi-4 Mini & -0.619** & -2.322** \\
     & \batteryitem $\times$ Phi-4 Mini & -1.318** & -1.183** \\
     & \sequentialitem $\times$ Qwen3 30B & -0.973** & -0.784* \\
     & \batteryitem $\times$ Qwen3 30B & -0.418** & -0.372 \\
     & \sequentialitem $\times$ Qwen3 4B & -1.125** & -0.885* \\
     & \batteryitem $\times$ Qwen3 4B & 0.490** & 1.574** \\
     & \sequentialitem $\times$ Gemma 3 12B & -1.218** & -1.366** \\
     & \batteryitem $\times$ Gemma 3 12B & -0.843** & -1.832** \\
     & \sequentialitem $\times$ Gemma 3 27B & -0.779** & -0.625 \\
     & \batteryitem $\times$ Gemma 3 27B & 0.087 & 0.250 \\     
     & \sequentialitem $\times$ Gemma 3 4B & 0.652** & 0.637 \\
     & \batteryitem $\times$ Gemma 3 4B & 2.175** & 2.394** \\

    \end{tblr}
    \caption{\textbf{Regression Results for MAE and Wasserstein Distance. ($\downarrow$)} 
    Model (1) uses OLS on Mean Absolute Error. Model (2) uses WLS on Wasserstein Distance, weighted by subpopulation count.
    Significance levels are based on Benjamini–Hochberg corrected p-values. We can see significant effects for both the questionnaire presentation, but also for the interaction between smaller models and the presentation.
    Reference categories: \textit{Presentation: \singleitem}, \textit{Model: Llama-3.3-70B-Instruct}.
    $\text{*}\,p<0.05,\ \text{**}\,p < 0.01$}.
    \label{tab:regression_results_questionnaire_pres}
\end{table*}

\section{Response Generation OLS Regressions} \label{app:response_ols}

We obtain the subpopulation-level alignment for each simulation specification and subpopulation, as described in Section~\ref{sec:ResponseGenerationMethod}. To identify significant differences in survey response alignment between the response generation methods, we fit the following OLS regression model separately on each dataset (see Table~\ref{tab:response_ols_long}): We use the per-subpopulation total variation distance $(\downarrow)$ as the dependent variable and Survey Response Generation Method (reference: Restricted Choice), LLM (reference: Llama 8B), and prompt perturbation (reference: Full Text response options) as independent variables.
We use cluster-robust SEs, clustering by seed × decoding strategy, which allows for arbitrary correlation and heteroskedasticity within clusters while assuming independence across clusters. This appropriately reflects the repeated-measures structure of our evaluation.
We do not include interaction terms into the OLS model to mitigate multicollinearity---all VIF values are < 3.
We apply Benjamini–Hochberg correction across all reported coefficients in all datasets.
Key coefficients for the Verbalized Distribution Method, as well as OLMo 32B and Qwen 32B remain significant even under Bonferroni correction, although Bonferroni is known to be overly conservative in regression settings with correlated predictors. 

\begin{table*}[p]
    \centering
    \small
    \begin{tblr}{
      colsep=8pt,
      colspec={Q Q X X X X},
      rowsep=1.5pt,
      width=\textwidth,
      column{1}={font=\bfseries},
      row{1}={font=\bfseries},
      hline{2,3,6,8,10,19}={-}{},
      hline{1,22}={-}{0.75pt},
      hborder{1,2,3,6,8,10,19,22}={belowspace=3pt}
    }
     &  & ANES 2016 & GLES 2017 & GLES 2025 & ATP 2021 \\
    Intercept & & 0.374** & 0.312** & 0.288** & 0.503** \\
    \SetCell[r=7]{h} {Response\\Generation Method} & {\textcolor{tab_blue}{\textbf{\rule{0.9ex}{0.9ex}}}}\;First-Token Probabilities & -0.003 & 0.147** & 0.194** & -0.049* \\
     & {\textcolor{tab_blue}{\textbf{\rule{0.9ex}{0.9ex}}}}\;First-Token Restricted & 0.064** & 0.220** & 0.234** & -0.005 \\
     & {\textcolor{tab_blue}{\textbf{\rule{0.9ex}{0.9ex}}}}\;Answer Prefix & -0.002 & 0.047* & 0.085** & -0.082** \\
     & {\textcolor{tab_blue2}{\textbf{\rule{0.9ex}{0.9ex}}}}\;Restricted Reasoning & 0.017 & -0.035* & -0.026 & -0.084** \\
     & {\textcolor{tab_blue2}{\textbf{\rule{0.9ex}{0.9ex}}}}\;Verbalized Distribution & -0.074** & -0.057** & -0.013 & -0.168** \\
     & {\textcolor{tab_blue3}{\textbf{\rule{0.9ex}{0.9ex}}}}\;Open-Ended Classif. & 0.026 & -0.011 & -0.027 & -0.051** \\
     & {\textcolor{tab_blue3}{\textbf{\rule{0.9ex}{0.9ex}}}}\;Open-Ended Distrib. & -0.006 & -0.052** & -0.037* & -0.082** \\
    \SetCell[r=9]{h} {Model} & Llama 3B & -0.051* & 0.031 & 0.066** & -0.039* \\
     & Llama 70B & -0.052* & -0.089** & -0.127** & 0.007 \\
     & OLMo 1B & -0.023 & 0.109** & 0.114** & 0.109** \\
     & OLMo 7B & -0.062** & 0.070** & 0.077** & -0.030 \\
     & OLMo 32B & -0.070** & -0.073** & -0.109** & 0.016 \\
     & Qwen 8B & 0.016 & 0.020 & -0.050* & 0.075** \\
     & Qwen 8B with Reasoning & -0.012 & 0.002 & -0.010 & 0.019 \\
     & Qwen 32B & -0.076** & -0.108** & -0.161** & -0.036* \\
     & Qwen 32B with Reasoning & -0.056** & -0.067** & -0.081* & -0.106** \\
    \SetCell[r=3]{h} {Response Option\\Variants} & Full Text, Reversed & 0.001 & -0.005 & 0.037 & -0.003 \\
     & Indexed & 0.010 & 0.003 & 0.000 & 0.022* \\
     & Indexed, Reversed & 0.035* & 0.011 & 0.026 & 0.030** \\
    \end{tblr}
    \caption{\textbf{Impact of {\color{tab_blue}$\blacksquare$}~Response Generation Methods on Subpopulation-Level Alignment $(\downarrow)$.}
    OLS regression coefficients by dataset with total variation distance $(\downarrow)$ as the dependent variable and Survey Response Generation Method (reference: Restricted Choice), LLM (reference: Llama 8B), and prompt perturbation (reference: Full Text response options) as independent variables. \textbf{The Verbalized Distribution Method and larger models lead to significant improvements.} $\text{*}\,p<0.05,\ \text{**}\,p < 0.01 
    \;\text{(Benjamini–Hochberg corrected)}$}
    \label{tab:response_ols_long}
\end{table*}

\end{document}